IAC-23-A3-3B

# Enabling In-Situ Resources Utilisation by leveraging collaborative robotics and astronaut-robot interaction

**Silvia Romero-Azpitarte[a], Cristina Luna[a]\*, Alba Guerra[a], Mercedes Alonso[a], Pablo Romeo Manrique[a], Marina L. Seoane[a], Daniel Olayo[a], Almudena Moreno[a], Pablo Castellanos[a], Fernando Gandía[a], Gianfranco Visentin[b]**

[a] *GMV Aerospace and Defence SAU, Calle de Isaac Newton 11, Tres Cantos, Madrid, Spain*, cluna@gmv.com.
[b] *ESTEC, ESA, Keplerlaan 1, 2201 AZ Noordwijk, The Netherlands.*
\* Corresponding Author

**Abstract**

Space exploration and establishing human presence on other planets demand advanced technology and effective collaboration between robots and astronauts. Efficient space resource utilization is also vital for extraterrestrial settlements. The Collaborative In-Situ Resources Utilisation (CISRU) project has developed a software suite comprising five key modules. The first module manages multi-agent autonomy, facilitating communication between agents and mission control. The second focuses on environment perception, employing AI algorithms for tasks like environment segmentation and object pose estimation. The third module ensures safe navigation, covering obstacle avoidance, social navigation with astronauts, and cooperation among robots. The fourth module addresses manipulation functions, including multi-tool capabilities and tool-changer design for diverse tasks in In-Situ Resources Utilization (ISRU) scenarios. Finally, the fifth module controls cooperative behaviour, incorporating astronaut commands, Mixed Reality interfaces, map fusion, task supervision, and error control.

The suite was tested using an astronaut-rover interaction dataset in a planetary environment and GMV SPoT analogue environments. Results demonstrate the advantages of E4 autonomy and AI in space systems, benefiting astronaut-robot collaboration. This paper details CISRU's development, field test preparation, and analysis, highlighting its potential to revolutionize planetary exploration through AI-powered technology.

**Keywords:** rover, ISRU, collaborative, artificial intelligence, manipulation

## 1. Introduction

Over the past few decades, there has been a clear international interest in human presence both on the Moon and Mars [1]. Achieving temporary and permanent human presence beyond our planet requires the development of a multitude of different technological systems, many of which are related to space robotics and automation.

In terms of robotics, astronauts will need to closely collaborate with various robots on the planetary surface. These collaborative tasks can range from regolith collection, In-Situ Resources Utilization (ISRU) activities, to habitat inspection and supervision. From this need arises the CISRU project (Collaborative ISRU), with the motivation to create an AI-enabled SW suite, regardless of the hardware platform used, able to support the programming of complex robot-robot and robot-human applications. It is mainly targeted to fuel-up the ISRU scenario for Moon and Mars exploration but also a wide application potential in terrestrial applications that can benefit from autonomy and collaborative work.

## 2. State-of-the-art and CISRU use cases

Collaborative ISRU refers to the utilization of local resources found on celestial bodies, such as the Moon or Mars, to support human and robotic missions. This approach aims to reduce the dependency on Earth for critical resources, such as water, oxygen, and building materials, making long-duration space exploration and colonization more sustainable [2].

In this context, there are already multiple projects whose objective has been to work to advance in this field:
- PERASPERA (Plan European Roadmap and Activities for Space Exploration and Robotic Activities) is an initiative by the European Commission to promote and coordinate research and development efforts related to robotic exploration and ISRU. It includes several key aspects such as technology development, resource prospecting and ISRU demonstrations [3]. ERGO is one of the projects of this plan with impact on CISRU [4].
- GOTCHA (GOAC TRL Increase convenience enhancements hardening and application extension) is ERGO'S predecessor. It focused on

developing a robotic autonomous controller for the space domain [5].
- ERGO (European Robotic Goal-Oriented Autonomous Controller): is part of the PERASPERA projects. It focuses specifically on the development of autonomous control systems for space robotics. It plays a significant role in collaborative ISRU by autonomous robotics, human-robot collaboration and robotic precursors [5]. It accomplished to develop a goal-oriented autonomy system suitable to be applied to different space robots operating in harsh environments, both space and terrestrial, achieving a 1.4 km autonomous long traverse [5].
- TASTE is a Model Driven Engineering approach that provides a methodology and tools to build dependable embedded software with real-time constraints. It has been developed as a follow-on of the ASSERT EC-FP6 project and is promoted by ESA.
- ROS 2 (Robot Operating System 2) is an open-source software development kit for robotics applications [6]. SpaceROS is an open-source space robotics framework for developing flight-quality robotic and autonomous space systems of recent creation [7] ROS2 is enabling current advances in collaborative robots, and the intention to create a standard for spatial purposes is very promising.

*2.1 Use-case 1: robot-robot collaboration*

The current state of the art in multi-robot collaboration represents a fascinating intersection of robotics, artificial intelligence, and automation technologies. Multi-robot collaboration is finding applications across various domains, including search and rescue, agriculture, manufacturing, and space exploration. These robots often need to work together to accomplish complex tasks efficiently.

Coordination mechanisms have become more sophisticated. Robots communicate through wireless networks, enabling them to share information about their surroundings, tasks, and status. Decentralized algorithms, such as consensus-based approaches and auction-based task allocation, allow robots to make collective decisions and allocate tasks optimally [8].

These robots are usually heterogenous in form or capabilities, to tackle a wider range of challenges. However, managing the heterogeneity presents an added difficulty in terms of coordination and communication.

One very interesting approach in multi-robotic systems, although not used in space yet is swarm robotics. Inspired by nature, swarm robotics involves large groups of relatively simple robots that collaborate based on local interactions. These systems are highly scalable and resilient, making them suitable for tasks like environmental monitoring and exploration [9].

In summary, the current challenges in multi-robot collaborative systems include efficient path planning in complex environments, ensuring robust communication in adverse conditions or handling dynamic task allocation among others. Current research frontiers focus on enhancing robot decision-making in uncertain environments and developing algorithms for large-scale multi-robot systems [8]. This can be especially helpful in the harsh and remote environments of Mars and the Moon, where multi-robotic systems can provide redundancy in case one robot encounters technical issues or fails. Backup robots can ensure that critical tasks can still be accomplished, reducing mission risks. Also, different robots within a multi-robotic system can be assigned to perform various tasks simultaneously. This parallelism enhances mission efficiency, allowing for multiple tasks to be completed concurrently, which is especially important in time-sensitive situations.

*2.2 Use-case 2: astronaut-robot collaboration*

The state of the art in astronaut-robot collaboration represents a pivotal aspect of space exploration and is characterized by several noteworthy developments. Robotic systems, such as the Canadarm and Canadarm2 on the Space Shuttle and the International Space Station (ISS), have been instrumental in tasks like capturing cargo spacecraft, performing repairs, and assisting with spacewalks [10].

On the other hand, humanoid robots, like Robonaut, Robonaut2 and CIMON (Crew Interactive Mobile Companion), have been designed to work alongside astronauts on the ISS [11], [12]. These robots have advanced mobility and dexterity, enabling them to perform tasks that were traditionally exclusive to human crew members.

Robot-astronaut collaboration plays a pivotal role in planetary exploration missions. Rovers can be used to scout ahead, map terrain, and identify potential hazards, allowing human astronauts to make more informed decisions about where to go and how to navigate. This collaboration optimizes exploration efficiency and minimizes risks. This was exampled by Curiosity and Perseverance, which conduct autonomous operations on Mars, collecting data and assisting in scientific research. Moreover, there is a current effort put in merging astronaut and robot capabilities in planetary settlements and exploration [13]. As space agencies plan for long-duration missions to the Moon, Mars, and beyond, astronaut-robot collaboration will continue to evolve.

The development of advanced robotic systems and AI-driven capabilities will be crucial for the success of these ambitious missions, given the limitation in resources and the precision required by the missions.

### 3. CISRU Software architecture

In this section, we describe the overall architecture of the developed software suite. The suite, as mentioned above, is divided into 5 different modules, that are designed to work simultaneously and to exchange different data to obtain a successful result.

The different interfaces of the modules were developed accordingly, and, in some cases, they make use of previous technologies such as ROS2 interface messages or ERGO Agent messages.

The following image shows how the different modules relate and the relationship between the hardware instances and the modules.

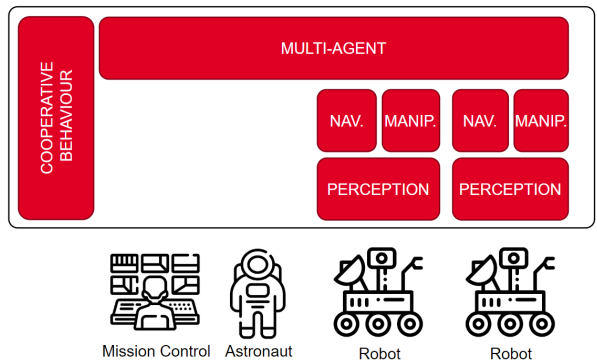

**Figure 1 CISRU software suite overview**

### 4.1 Multi-agent components

Based on the ERGO architecture, in which the agent is responsible for the correct interaction among the different reactors and the overall Agent behaviour. For the communication with the reactors, the agent controller uses a specific interface based on goals and observations.

In a multi-agent system, two or more agents interact with each other to accomplish more complex tasks.

Using a specific reactor, the MAS (Multi-Agent Synchronization) reactor, the different agent instances can send and receive goals and observations from the other agents.

The agents can work in different autonomy levels. This autonomy levels follow the ECSS standard [14] and determine what telecommands are processed and the level of autonomous decision taking enabled. Since CISRU is a goal-oriented mission with the ability of re-planning, the autonomy level acquired is E4. Only E4-level goals are accepted in this level, by decomposing, planning and scheduling goals, the system is able to execute goal-oriented mission operations on-board. If the execution deviates from the expected, the system can adapt the plan accordingly.

The designed architecture is composed of two Agents: Leader and Secondary. In the Human-Robot use case, the Leader will be the robot helping the astronaut with the tasks. In the Robot-Robot use case, the Leader oversees managing the mission, while the Secondary offers support.

The goals decomposition represents the expected execution inside the agent for each high-level goal received. This decomposition is relevant in the E4 autonomy level. In other levels, the operator or the astronaut has full control of the system. This decomposition is divided into the two different use cases:
- Human-Robot represents the collaboration between the Leader robot and the astronaut. The secondary robot is not needed for these actions. The first goal to execute is the inspection solar panels at a set of points. Meanwhile, the Leader will supervise and alert in case of an emergency is observed.
- Robot-Robot represents the collaboration between two robots to map an area and analyse potential samples to be taken to the base. Both robots are used in this scenario collaboratively. The main objective consists of mapping an area and analysing certain points with the Leader robot. If the Leader finds an interesting sample, it will call the secondary to store it. After the sample has been stored, the secondary robot can either take it to the base or continue the mapping of its area. When taking the sample to the base, an astronaut will empty the storage and inform the robot.

### 4.2 Perception components

The perception components oversee the vision and perception of the robots. They include Human-Machine interaction detection, equipment anomaly detection, and Human emergency situations. These models are low resource consumption and run in a MyriadX VPU, a computing unit that is being tested in space environments as a possibility to process AI models in extreme conditions [15]. The specific architecture used is Mobilenet-ssd [16] but trained for different targets. This architecture uses Single Shot Detection as one of the key features for its fast and quite precise output [17], [18], and was selected because of the input data received (a continuous video) that permitted us to have a less precise result that could be corrected over time thanks to the fast predictions of the model.

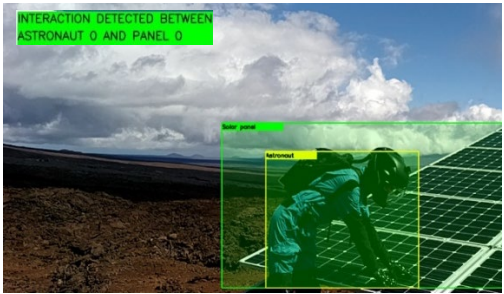

**Figure 2 Interaction detection between astronaut and solar panel and identification of both instances at HI-SEAS Martian analogue environment. Detection made by Mobilenet-SSD.**

Semantic segmentation is also implemented as a 2D semantic segmentation model able to enhance the 3D map created by the stereo cameras of the rovers. This creates a 3D Semantic Map filtered using the 2D model, the point cloud and the depth calculator. As output, the data produced by the perception components are used in the cooperative behaviour, navigation, and manipulation components as well as in the visualisation tool. The model used for this specific task could not produce much noise because it would cause sensible damage on the mapping. Therefore, DeepLabV3+ [19] was selected and retrained with the appropriate datasets to produce the masks that later filter the point-cloud. This architecture has already been tested for similar purposes and the results are very promising [20].

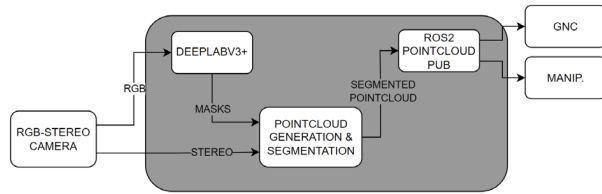

**Figure 3 Semantic segmentation of the perception component.**

As shown in Figure 3, the RGB image is processed to obtain the masks and the result is used to filter the point cloud, as shown in Figure 4.

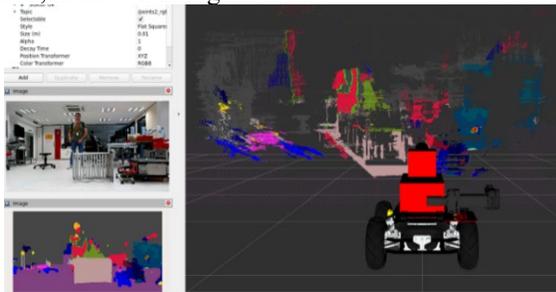

**Figure 4 Semantic segmentation of point-cloud, visualized in RVIZ. It shows the result for an indoor environment.**

To train the solar-panel error detection the datasets

The data used to train all these models but solar panels one, was obtained at ESTEC's dedicated planetary environment, in addition to different simulators and testing in the analogue environment, GMV SPoT and HI-SEAS Martian analogue environment in Hawaii. The identification dataset (where labels are bounding boxes) was created by pseudo-automating the process and training specific neural networks to automatically label a set of more than 2000 images. This dataset was used to train interaction detection and astronaut monitoring.

These identification labels were then used to generate the masks for semantic segmentation, making use of SAM [21] and merging it with other segmentation datasets [22]. The result for preparing the segmentation dataset was as shown in Figure 5.

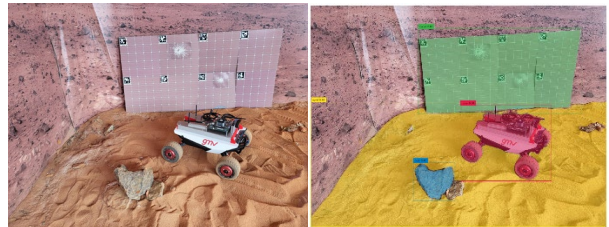

**Figure 5 Auto-generated semantic segmented masks to train DeepLabV3+.**

The final generated dataset incorporates the following labels: astronaut, rock, rover, solar panel, regolith/Martian floor. This dataset was finally used to train DeepLabV3+ for the segmentation of the environment and thus the enhancement of the GNC system.

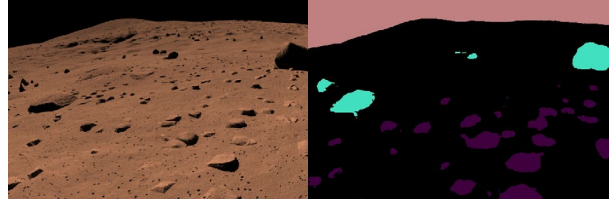

**Figure 6 Result of semantic segmentation of the Martian environment done by trained DeepLabV3+.**

*4.3 Guidance, navigation and control components*

The Navigation is the first necessary component to have GNC system for each robot to make them autonomous and to be able to execute the different task on each use case. This component can be divided in two parts: a global localisation and mapping based on Visual SLAM (Simultaneous Localisation and Mapping) together with the odometry information and a path-planning module and trajectory control based on Fast Marching Square [23].

The SLAM system is based on stereo vision, facing the corresponding challenges of the sensor. The map is then processed using NAV2 and the planification algorithm is implemented as a NAV2 plugin. This resulted in a very modular and understandable solution that can be easily reused.

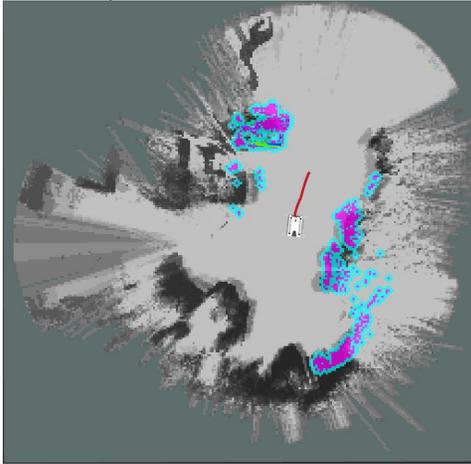

**Figure 7 Navigation on a SLAM map using NAV2 plugins.**

*4.4 Manipulation components*

The manipulation component handles the control and planning of the robotic arm, as well as the procedure of the sample collection algorithm. It is responsible for low-level procedures, such as commanding the motors of the manipulator, obtain the position of each link of the robotic arm and modify the speed or acceleration of the different parts of the arm. It also incorporates high-level actions including motion planning and conducting each of the necessary steps for tool changing and sample collection.

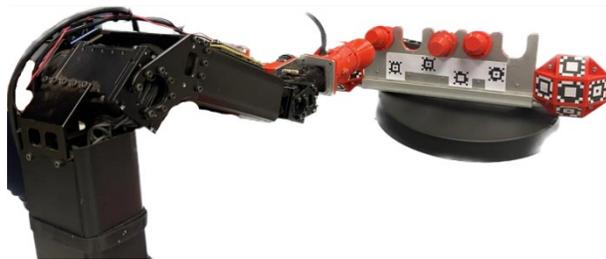

**Figure 8 Manipulation system. Arm with tool loaded and tool changer.**

The tool-changer component is used to mount/dismount the tool from the robotic arm, situated on a specific slot mounted on the secondary rover. It has a manager to control the arm and a state machine to safely control the tool-changer. A vision algorithm is integrated to obtain a precise position of the tool relative to the end effector of the robotic arm. Given such position, the system can calculate the trajectory needed to catch the tool or in case of unreachability the cooperative module is informed.

Once a tool is mounted on the end of the robotic arm, it may be commanded to recollect and store a soil sample. The sample collection component is responsible for obtaining a sample of the ground -after checking the tool is correctly assembled- and to move the sample to the correct storage slot and unload it from the robotic arm.

*4.5 Cooperative behaviour components*

The cooperative behaviour module is a very complex and heterogenous software component, that deals with functions as different as Mixed Reality interfaces (for the enhancement of the communications with the astronaut) and map fusion among others. The key factor of this module is the synchronization and orchestration of the communications of the different data used in each of the functionalities.

The Mixed Reality interfaces were developed by integrating some of the ROS2 message interfaces with the Hololens2 available software. This ensured that the accessible design of some of the glasses' interfaces could be reused, as well as allowing us to focus on the appropriate information and representation to send to the astronaut. In this context, it was very important to observe the accessibility and avoid the overexposure to information that might result in an inability to focus on the tasks.

The Map Fusion component enables robots to gather information about their global positions within the environment and their relative positions to each other. The purpose of having a shared map is to establish a common scenario for the robots, facilitating robot-to-robot interaction. As a result, one robot can transmit instructions to the other regarding coordinates for movement or provide information about the environment in specific locations. To create a comprehensive map, both robots must have mapped the same environment, where similar features can be detected. Key-points are detected using feature detection descriptors and matched using a Brute-force matcher. The resultant matches are used to enable pixel-level image fusion.

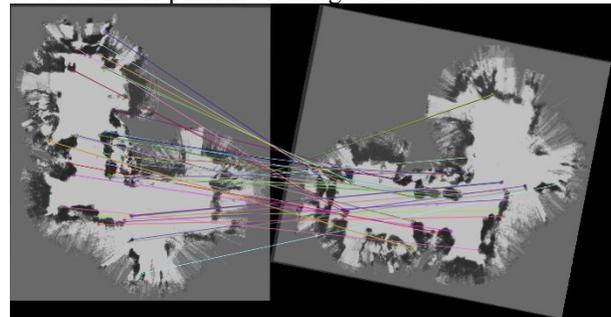

**Figure 9 Map fusion by key-point detection and matching in different orientations and different levels of completeness of the same map.**

The tasks of supervision, emergency and error control are implemented as a parallel thread that overlooks the system and reports the corresponding information to different actors depending on the event.

The errors and sudden problems are reported to the Mission Control (in case an error cannot be handled)

In case of emergency (such as astronaut falling) it first asks the astronaut to make sure that he/she is safe and whether there is a real emergency. If no answer is received by the system, the Mission Control is the alerted.

In case of supervision, the thread controls that the astronauts are doing what they are supposed to do (in the practical case astronaut A was assigned the maintenance of solar panel 1 and if the astronaut worked on solar panel 2 the astronaut and the Mission Control were alerted). This feature also implemented instance tracking for both astronauts and robots.

**4. Test campaign**

The test campaign has been designed as an incremental plan, in which each component was tested on its own first inside the lab and then in the analogue environment, GMV SPoT. After passing these tests, the modules are integrated into the decided hardware specifications and put to work simultaneously and collaboratively.

*4.1 Test preparation*

The preparation for testing the CISRU suite's capabilities involved the creation of an astronaut-rover interaction dataset in a dedicated planetary environment, replicating the challenges and conditions of space exploration. Additionally, we leveraged various simulators to simulate a range of space exploration scenarios and subjected the suite to testing in the analogue environment of GMV SPoT.

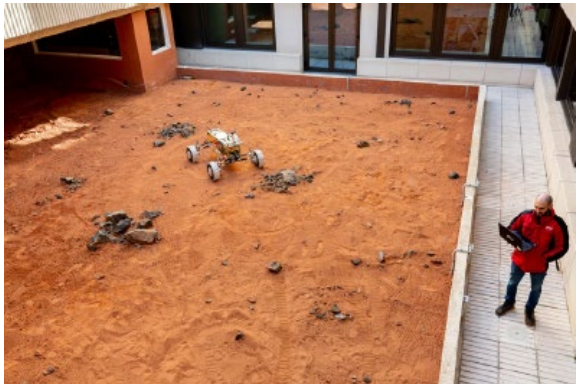
**Figure 10 GMV SPoT.**

The robots prepared and used for the test are our Lunar and Martian Autonomous Reconnaissance Rover (LAMARR) and our Mini Autonomous Explorer (MAE). LAMARR is acting as the leader robot and MAE as the secondary robot while both are able to swap their roles.

LAMARR is equipped with a ROBOTIS Manipulator-H which carries the tool-changer, a VisNIR camera to detect interesting points on the soil and the navigation and localisation sensors (IMU, stereo cameras, etc). MAE is also equipped with the navigation and localisation sensors as well as a Human Machine Interface (HMI) and the tools for the leader robot (different shovels and brushes).

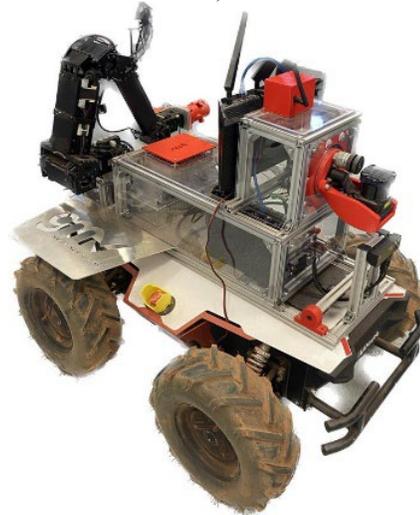
**Figure 11 Leader robot LAMARR.**

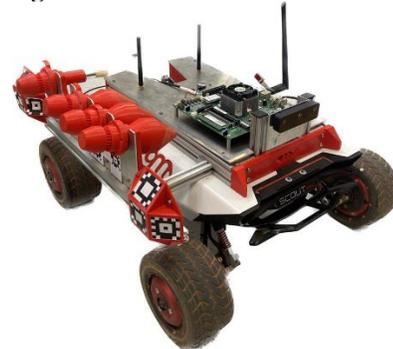
**Figure 12 Secondary robot MAE.**

The GMV SPoT facility is extremely useful since we can test our algorithms in a Martian-like scenario as well as test the whole performance of the software suite outdoors. In addition, we placed a mock-up of a solar panel array so the robots and the astronaut can collaborate to achieve the goals of the Use Case 2.

Previously to the test execution, a test plan was drafted, in which a set of tests for each component was designed. The test plan for the collaboration component was scheduled last, since this component in a sense implies the integration and correct working of the perception, navigation, manipulation, and multi-agent components.

*4.2 Test execution*

Throughout these tests, a meticulous assessment of the suite's performance unfolded, focusing on its capacity to comprehend, navigate, and engage with the surrounding environment and its various agents. The outcomes of these assessments reaffirmed the suite's potential in facilitating effective astronaut-robot collaboration while substantiating the merits of the E4 autonomy level and the AI capabilities within space systems.

The following key points describe the high-level tests executed for every component:
- The system can identify a crack on a solar panel like surface and report where it was found.
  The system can detect and track astronaut instances and detect when any of these instances is fallen. It can then report it.
  The system and detect and track solar panels and rover instances and detect when any astronaut instance is within the interactable region of any of these objects. The system can then report it.
- The system can map a region and avoid obstacles. The map is filtered using semantic segmentation to enhance the point cloud originated from the stereo camera. The system can autonomously plan and navigate through the map to accomplish the missions commanded.
- The system can manipulate the tool exchanger device to take soil samples in the desired sites. Both robots collaborate to fulfil this requirement.
- The system can coordinate the execution of both robots and the astronaut to perform high-level tasks.
- The system is able to coordinate the mapping between the robots (map fusion), as well as collaborate with the astronaut, making use of the mixed reality device selected for the mission. The system can also track the malfunctioning on any component and report them.

This rigorous testing regimen underscored the transformative promise of the CISRU suite for future planetary exploration endeavours.

**5. Results**

The results are examined by analysing each module and then the overall functionality of the suite.

In terms of perception and navigation, the results are more favourable in an environment where obstacles are present, as it is imperative to have reference points for generating a point cloud and accurately marking the locations of obstacles. In the laboratory setting, there are no challenges, given the abundance of objects. Conversely, within the GMV SPoT, rocks prove valuable in detecting point clouds at specific depths, and where there are no rocks, the map is marked as unknown.

The part of perception dealing with astronaut and environment monitoring was very successful for Martian-like terrains. Astronauts and robots were detected and tracked even when the whole body was not on display. On the counter part, this is a non-exact technique so a new retrain and dataset would be most likely needed for Lunar-like terrains. A new subset of the dataset of detection was required to adapt the model for indoor facilities such as the lab where the project was developed.

In terms of planning, the robots generate a path even when the environment has not been explored yet. If the path cannot be established due to the presence of an obstacle, it generates a new one. This process continues iteratively until there are no more available options, indicating that the goal is unreachable. In this sense, we consider that this is rather useful, given that the spatial environment is usually unknown and if no path can be stablished then no movement would be achieved.

Manipulation, as mentioned above, is a fundamental part of most ISRU projects. Our proposition included new tools developed for different manipulation of the possible resource of Mars and Moon surfaces. The new tools proved to be very appropriate for the tasks that they were designed for, and the arm located on the Leader robot was able to make use of this tools autonomously to perform the missions and in the cases where the tools were out of reach the system was able to report it correctly. It has been proven that the development of more tools is a challenge for the future of spatial exploration.

The multi-agent module has shown a great performance in both use cases that were scheduled in CISRU. As part of the multi-agent functionalities, a wearable computer with touchable screen was designed for the astronaut, to include him/her in the Agent system of communications. The network analysis showed that the processes and communications deployed were not too heavy.

The overall collaborative behaviour and the added module on top increased the computing load, although the result was acceptable, and it showed that collaboration between robots and astronaut-robots is not only necessary but also possible.

**6. Discussion**

The CISRU project represents a significant milestone in the realm of space exploration and robotics, marked by its innovative integration of various techniques to enhance the capabilities of multi-agent robotic systems. One of the project's notable achievements is the successful incorporation of cutting-edge technologies, such as artificial intelligence for perception and map merging. While the concept of AI-enhanced robotics for extraterrestrial exploration is not entirely new, the novel implementation of these techniques in the project has demonstrated remarkable adaptability and efficiency in navigating the challenges of Martian-like environments.

However, it is essential to acknowledge the project's limitations. The experiments were primarily conducted in controlled Martian-like environments, and extrapolating these findings to the harsh and unpredictable realities of lunar or Martian terrain remains a significant challenge. Additionally, the project's communication infrastructure, though functional, is constrained by limited bandwidth and range, which could pose challenges during extended missions or in more remote areas.

Another limitation is the current multi-agent system's size, comprising only two robots. This limitation impacts the project's scalability and the complexity of collaborative tasks, particularly in scenarios requiring intricate manipulation or coordinated actions between multiple robots. Notably, manipulation tasks involving the collaboration of both robots were found to be challenging and could benefit from further development.

Nonetheless, the project boasts a modular and hardware-independent codebase, a substantial achievement that enhances its adaptability to different robotic platforms and hardware configurations. This flexibility not only simplifies hardware upgrades but also streamlines mission planning and execution.

In looking to the future, several avenues for improvement and expansion present themselves. Increasing the number of robots and astronauts collaborating within the system could enhance mission efficiency and versatility, enabling a more extensive range of tasks and experiments. Moreover, the development of a more accessible interface for astronauts would facilitate smoother communication and control, allowing them to leverage the capabilities of the robotic team effectively.

To address the complexities of manipulation tasks, the project could benefit from the inclusion of additional tools and improved manipulation capabilities, allowing robots to perform a wider range of tasks autonomously. Furthermore, enhancing the AI system with more extensive training and learning could boost the robots' decision-making capabilities, making them better equipped to adapt to unforeseen challenges.

Finally, efforts to reduce resource consumption in the Guidance, Navigation, and Control (GNC) system could lead to more efficient operations, conserving power, and resources for extended missions. In summary, while the collaborative ISRU project has achieved remarkable milestones, its ongoing evolution holds promise for further advancing our understanding of robotic collaboration in extraterrestrial exploration.

## 7. Conclusions

In conclusion, the collaborative ISRU model presented in this paper represents a significant leap forward in our approach to extraterrestrial resource utilization. The successful teamwork between two robots and an astronaut has demonstrated the feasibility and efficiency of such collaborations in harsh planetary environments. These findings underline the potential for optimizing resource utilization, reducing mission risks, and expanding the scope of human exploration beyond Earth.

As we look to future missions on the Moon, Mars, and beyond, this collaborative paradigm offers a promising blueprint for sustainable and productive space exploration. It also calls for ongoing research and development efforts to refine the capabilities and autonomy of robotic systems, ensuring they continue to complement and enhance the capabilities of human astronauts in the quest to unlock the secrets of the universe.

**Acknowledgements**
The CISRU project has been partially founded by both ESA and GMV under grant agreement No. 4000135391/21/NL/GLC/ZK.